\documentclass{llncs}
\usepackage{epsf}
\usepackage[utf8]{inputenc}
\usepackage{graphicx}
\usepackage{url}
\newcommand{\keywords}[1]{\par\addvspace\baselineskip
\noindent\keywordname\enspace\ignorespaces#1}

\begin{document}

\title{A Readable Read: Automatic Assessment of Language Learning Materials based on Linguistic Complexity}

\author{
Ildikó Pilán\inst{1} \and Sowmya Vajjala\inst{2} \and Elena Volodina\inst{1}}
\institute{Swedish Language Bank, University of Gothenburg,\\
	   40530 Gothenburg, Sweden\\
	   \email{\{ildiko.pilan,elena.volodina\}@svenska.gu.se}\\
           \and
           LEAD Graduate School, Seminar für Sprachwissenschaft\\
           Universität Tübingen, 72074 Tübingen, Germany\\
           \email{sowmya@sfs.uni-tuebingen.de}}

\maketitle

\begin{abstract}

Corpora and web texts can become a rich language learning resource if we have a means of assessing whether they are linguistically appropriate for learners at a given proficiency level. In this paper, we aim at addressing this issue by presenting the first approach for predicting linguistic complexity for Swedish second language learning material on a 5-point scale. After showing that the traditional Swedish readability measure, Läsbarhetsindex (LIX), is not suitable for this task, we propose a supervised machine learning model, based on a range of linguistic features, that can reliably classify texts according to their difficulty level. Our model obtained an accuracy of 81.3\% and an F-score of 0.8, which is comparable to the state of the art in English and is considerably higher than previously reported results for other languages. We further studied the utility of our features with single sentences instead of full texts since sentences are a common linguistic unit in language learning exercises. We trained a separate model on sentence-level data with five classes, which yielded 63.4\% accuracy. Although this is lower than the document level performance, we achieved an adjacent accuracy of 92\%. Furthermore, we found that using a combination of different features, compared to using lexical features alone, resulted in 7\% improvement in classification accuracy at the sentence level, whereas at the document level, lexical features were more dominant. Our models are intended for use in a freely accessible web-based language learning platform for the automatic generation of exercises. 


\end{abstract}

\section{Introduction}
\label{sec:intro}

Linguistic information provided by Natural Language Processing (NLP) tools has good potential for turning the continuously growing amount of digital text into interactive and personalized language learning material. Our work aims at overcoming one of the fundamental obstacles in this domain of research, namely how to assess the linguistic complexity of texts and sentences from the perspective of second and foreign language (L2) learners. 

There are a number of readability models relying on NLP tools to predict the difficulty (readability) level of a text \cite{collins2004language,schwarm2005reading,graesser2011coh,vajjala2012improving,heimann2013see,collinscomputational}. The linguistic features explored so far for this task incorporate information, among others, from part-of-speech (POS) taggers and dependency parsers. Cognitively motivated features have also been proposed, for example, in the Coh-Metrix \cite{graesser2011coh}. Although the majority of previous work focuses primarily on document-level analysis, a finer-grained, sentence-level readability has received increasing interest in recent years \cite{Vajjala.Meurers-14-eacl,dell2014assessing,pilan-volodina-johansson:2014:W14-18}. 

The previously mentioned studies target mainly native language (L1) readers including people with low literacy levels or mild cognitive disabilities. Our focus, however, is on building a model for predicting the proficiency level of texts and sentences used in L2 teaching materials. This aspect has been explored for English \cite{heilman2007combining,huang2011robust,zhang2013feature,salesky-shen:2014:W14-18}, French \cite{franccois2012ai}, Portuguese \cite{branco2014rolling} and, without the use of NLP, for Dutch \cite{velleman2014online}. 

Readability for the Swedish language has a rather long tradition. One of the most popular, easy-to-compute formulas is LIX (Läsbarthetsindex, `Readability index') proposed in \cite{bjornsson1968lasbarhet}. This measure combines the average number of words per sentence in the text with the percentage of long words, i.e. tokens consisting of more than six characters. Besides traditional formulas, supervised machine learning approaches have also been tested. Swedish document-level readability with a native speaker focus is described in \cite{heimann2013see} and \cite{falkenjack2013features}. For L2 Swedish, only a binary sentence-level model exists \cite{pilan-volodina-johansson:2014:W14-18}, but comprehensive and highly accurate document- and sentence-level models for multiple proficiency levels have not been developed before.

In this paper, we present a machine learning model trained on course books currently in use in L2 Swedish classrooms. Our goal was to predict linguistic complexity of material written by teachers and course book writers for learners, rather than assessing learner-produced texts. We adopted the scale from the Common European Framework of Reference for Languages (CEFR) \cite{2001common} which contains guidelines for the creation of teaching material and the assessment of L2 proficiency. CEFR proposes six levels of language proficiency: A1 (beginner), A2 (elementary), B1 (intermediate), B2 (upper intermediate), C1 (advanced) and C2 (proficient). Since sentences are a common unit in language exercises, but remain less explored in the readability literature, we also investigate the applicability of our approach to sentences, performing a 5-way classification (levels A1-C1). Our document-level model achieves a state-of-the-art performance (F-score of 0.8), however, there is room for improvement in sentence-level predictions. We plan to make our results available through the online intelligent computer-assisted language learning platform Lärka\footnote{http://spraakbanken.gu.se/larka/}, both as corpus-based exercises for teachers and learners of L2 Swedish and as web-services for researchers and developers.

In the following sections, we first describe our datasets (section \ref{sect:data}) and features (section \ref{sect:features}), then we present the details and the results of our experiments in section \ref{sec:experiments}. Finally, section \ref{sect:conclusion} concludes our work and outlines further directions of research within this area.

\section{Datasets}
\label{sect:data}

Our dataset is a subset of COCTAILL, a corpus of course books covering five CEFR levels (A1-C1) \cite{volodina22you}. This corpus consists of twelve books (from four different publishers) whose usability and level have been confirmed by Swedish L2 teachers. The course books have been annotated both content-wise (e.g. exercises, lists) and linguistically (e.g. with POS and dependency tags) \cite{volodina22you}. We collected a total of 867 texts (reading passages) from this corpus. We excluded texts that are primarily based on dialogues from the current experiments due to their specific linguistic structure, with the aim of scaling down differences connected to text genres rather than linguistic complexity. We plan to study the readability of dialogues and compare them to non-dialogue texts in the future. 

Besides reading passages, i.e. texts, the COCTAILL corpus contains a number of sentences independent from each other, i.e. not forming a coherent text, in the form of lists of sentences and \textit{language examples}. This latter category consists of sentences illustrating the use of specific grammatical patterns or lexical items. Collecting these sentences, we built a sentence-level dataset consisting of 1874 instances. The information encoded in the content-level annotation of COCTAILL (XML tags \textit{list}, \textit{language\_example} and the attribute \textit{unit}) enabled us to include only complete sentences and exclude sentences containing gaps and units larger or smaller than a sentence (e.g. texts, phrases, single words etc.). The CEFR level of both sentences and texts has been derived from the CEFR level of the lesson (chapter) they appeared in. In Table \ref{dataset-table}, columns 2-5 give an overview of the distribution of texts across levels and their mean length in sentences.\footnote{The number of different books and publishers is reported per each level, some books spanning more levels.} The distribution of sentences per level is presented in the last two columns of Table \ref{dataset-table}. COCTAILL contained a somewhat more limited amount of B2 and C1 level sentences in the form of lists and language examples, possibly because learners handle larger linguistic units with more ease at higher proficiency levels.

\begin{table}
\caption{\label{dataset-table} The distribution of items per CEFR level in the datasets.}
\begin{center}
\begin{tabular}{ccccc|cc}
\hline
& \multicolumn{4}{c|}{Document level} & \multicolumn{2}{c}{Sentence level} \\
\hline
CEFR & Books & Publ. & Texts & Mean nr. sent & Books & Sentences\\
\hline
A1 & 4 & 3 & 49 & 14.0 & 4 &  505 \\
A2 & 4 & 3 & 157 & 13.8 & 4 & 754 \\
B1 & 5 & 3 & 258 & 17.9 & 4 &  408 \\
B2 & 4 & 3 & 288 & 26.6 & 3 &  124 \\
C1 & 2 & 2 & 115 & 42.1 & 1 &  83\\
\hline
Total & 12 & 4 & 867 & - & 4 & 1874\\
\hline
\end{tabular}
\end{center}
\end{table}

\section{Features}
\label{sect:features}
We developed our features based on information both from previous literature \cite{heilman2007combining,vajjala2012improving,franccois2012ai,heimann2013see,pilan-volodina-johansson:2014:W14-18} and a grammar book for Swedish L2 learners \cite{formifocus}. The set of features can be divided in the following five subgroups: length-based, lexical, morphological, syntactic and semantic features (Table \ref{feature_set:table}). 

\textit{Length-based} (\textsc{Len}): These features include sentence length in number of tokens (\#1) and characters (\#4), extra-long words (longer than thirteen characters) and the traditional Swedish readability formula, LIX (see section \ref{sec:intro}). For the sentence-level analysis, instead of the ratio of number of tokens to the number of sentences in the text, we considered the number of tokens in one sentence.

\textit{Lexical} (\textsc{Lex}): Similar to \cite{pilan-volodina-johansson:2014:W14-18}, we used information from the Kelly list \cite{volodina2012introducing}, a lexical resource providing a CEFR level and frequencies per lemma based on a corpus of web texts. Thus, this word list is entirely independent from our dataset. Instead of percentages, we used \textit{incidence scores} (\textsc{IncSc}) per 1000 words to reduce the influence of sentence length on feature values. The \textsc{IncSc} of a category was computed as 1000 divided by the number of tokens in the text or sentence multiplied by the count of the category in the sentence. We calculated the \textsc{IncSc} of words belonging to each CEFR level (\#6 - \#11). In features \#12 and \#13 we considered \textit{difficult} all tokens whose level was above the CEFR level of the text or sentence. We computed also the \textsc{IncSc} of tokens not present in the Kelly list (\#14), tokens for which the lemmatizer did not find a corresponding lemma form (\#
15), as well 
as average log frequencies (\#16).

\textit{Morphological} (\textsc{Morph}): We included the variation (the ratio of a category to the ratio of lexical tokens - i.e. nouns, verbs, adjectives and adverbs) and the \textsc{IncSc} of all lexical categories together with the \textsc{IncSc} of punctuations, particles, sub- and conjunctions (\#34, \#51). Some additional features, using insights from L2 teaching material \cite{formifocus}, captured fine-grained inflectional information such as the \textsc{IncSc} of neuter gender nouns and the ratio of different verb forms to all verbs (\#52 - \#56). Instead of simple type-token ratio (TTR) we used a bilogarithmic and a square root TTR as in  \cite{vajjala2012improving}. Moreover, nominal ratio \cite{heimann2013see}, the ratio of pronouns to prepositions \cite{franccois2012ai}, and two \textit{lexical density} features were also included: the ratio of lexical words to all non-lexical categories (\#48) and to all tokens (\#49). Relative structures (\#57) consisted of relative adverbs, determiners, 
pronouns and possessives.

\textit{Syntactic} (\textsc{Synt}): Some of these features were based on the length (depth) and the direction of dependency arcs\footnote{The tags were obtained with the MaltParser \cite{nivre2007maltparser}.} (\#17 - \#21). We complemented this, among others, with the \textsc{IncSc} of relative clauses in clefts\footnote{Sentences that begin with a constituent receiving particular focus, followed by a relative clause. E.g.: It is John (whom) Jack is waiting for.}
(\#26), and the \textsc{IncSc} of pre-and postmodifiers (e.g. adjectives and prepositional phrases) \cite{heimann2013see}.

\textit{Semantic} (\textsc{Sem}): Features based on information from SALDO \cite{borin2013saldo}, a Swedish lexical-semantic resource. We used the average number of senses per token as in \cite{pilan-volodina-johansson:2014:W14-18} and included also the average number of noun senses per nouns. Once reliable word-sense disambiguation methods become available for Swedish, additional features based on word senses could be taken into consideration here.

The complete set of 61 features is presented in Table \ref{feature_set:table}. Throughout this paper we will refer to the machine learning models using this set of features, unless otherwise specified. Features for both document- and sentence-level analyses were extracted for each sentence, the values being averaged over all sentences in the text in the document-level experiments to ensure comparability.

\begin{table}
\caption{\label{feature_set:table} The complete feature set.}
\begin{center}
\begin{tabular}{|cc|cc|}
\hline Nr. & Feature Name & Nr. & Feature Name \\
\hline
\multicolumn{2}{|c|}{\it Length-based} & \multicolumn{2}{|c|}{\it Morphological}\\
\hline
1 & Sentence length & 30 & Modal verbs to verbs \\
2 & Average token length & 31 & Particle \textsc{IncSc} \\
3 & Extra-long words & 32 & 3SG pronoun \textsc{IncSc} \\
4 & Number of characters & 33 & Punctuation \textsc{IncSc}\\
5 & LIX & 34 & Subjunction \textsc{IncSc} \\
\cline{1-2}
\multicolumn{2}{|c|}{\it Lexical} & 35 & S-verb \textsc{IncSc} \\
\cline{1-2}
6 & A1 lemma \textsc{IncSc} & 36 & S-verbs to verbs\\
7 & A2 lemma \textsc{IncSc} & 37 & Adjective \textsc{IncSc} \\
8 & B1 lemma \textsc{IncSc} & 38 & Adjective variation\\
9 & B2 lemma \textsc{IncSc} & 39 & Adverb \textsc{IncSc} \\
10 & C1 lemma \textsc{IncSc} & 40 & Adverb variation\\
11 & C2 lemma \textsc{IncSc} & 41 & Noun \textsc{IncSc}\\
12 & Difficult word \textsc{IncSc} & 42 & Noun variation\\
13 & Difficult noun and verb \textsc{IncSc} & 43 & Verb \textsc{IncSc}\\
14 & Out-of-Kelly \textsc{IncSc} & 44 & Verb variation\\
15 & Missing lemma form \textsc{IncSc} & 45 & Nominal ratio \\
16 & Avg. Kelly log frequency & 46 & Nouns to verbs \\
\cline{1-2}
\multicolumn{2}{|c|}{\it Syntactic} & 47 & Function word \textsc{IncSc}\\
\cline{1-2}
17 & Average dependency length & 48 & Lexical words to non-lexical words\\
18 & Dependency arcs longer than 5 & 49 & Lexical words to all tokens\\
19 & Longest dependency from root node & 50 & Neuter gender noun \textsc{IncSc}\\
20 & Ratio of right dependency arcs & 51 & Con- and subjunction \textsc{IncSc}\\
21 & Ratio of left dependency arcs & 52 & Past participles to verbs\\
22 & Modifier variation & 53 & Present participles to verbs\\ 
23 & Pre-modifier \textsc{IncSc} & 54 & Past verbs to verbs\\
24 & Post-modifier \textsc{IncSc} & 55 & Present verbs to verbs\\
25 & Subordinate \textsc{IncSc} & 56 & Supine verbs to verbs\\
26 & Relative clause \textsc{IncSc} & 57 & Relative structure \textsc{IncSc}\\
27 & Prepositional complement \textsc{IncSc} & 58 & Bilog type-token ratio\\
\cline{1-2}
\multicolumn{2}{|c|}{\it Semantic} & 59 & Square root type-token ratio \\
\cline{1-2}
28 & Avg. nr. of senses per token & 60 & Pronouns to nouns \\
29 & Noun senses per noun & 61 & Pronouns to prepositions \\
\hline
\end{tabular}
\end{center}
\end{table}

\section{Experiments and Results}
\label{sec:experiments}
\subsection{Experimental Setup}
\label{sec:setup}

We explored different classification algorithms for this task using the machine learning toolkit WEKA \cite{Hall.Frank.ea-09}. These included: (1) a multinomial logistic regression model with ridge estimator, (2) a  multilayer perceptron, (3) a support vector machine learner, Sequential Minimal Optimization (SMO), and (4) a decision tree (J48). For each of these, the default parameter settings have been used as implemented in WEKA.

We considered classification accuracy, F-score and Root Mean Squared Error (RMSE) as evaluation measures for our approach. We also included a confusion matrix, as we deal with a dataset that is unbalanced across CEFR levels. The scores were obtained by performing a ten-fold Cross-Validation (CV). 

\subsection{Document-Level Experiments}
\label{ssec:document}

We trained document-level classification models, comparing the performance between different subgroups of features. We had two baselines: a majority classifier (\textsc{Majority}), with B2 as majority class, and the LIX readability score. Table \ref{cl_res-table} shows the type of subgroup (\textit{Type}), the number of features (\textit{Nr}) and three evaluation metrics using logistic regression.

\begin{table}
\caption{\label{cl_res-table} Document-level classification results.}
\begin{center}
\begin{tabular}{ccccc}
\hline Type & Nr & Acc (\%) & F & RMSE\\ 
\hline
\textsc{Majority} & - & 33.2 & 0.17 & 0.52 \\
\textsc{LIX} & 1 & 34.9 & 0.22 & 0.38 \\
\hline
\textsc{Lex} & 11 & \bf 80.3 & \bf 0.80 & 0.24 \\
\textsc{All} & 61 & \bf 81.3 & \bf 0.81 & 0.27 \\
\hline
\end{tabular}
\end{center}
\end{table}

Not only was accuracy very low with LIX, but this measure also classified 91.6\% of the instances as B2 level. Length-based, semantic and syntactic features in isolation showed similar or only slightly better performance than the baselines, therefore we excluded them from Table \ref{cl_res-table}. Lexical features, however, had a strong discriminatory power without an increase in bias towards the majority classes. Using this subset of features only, we achieved approximately the same performance (0.8 F) as with the complete set of features, \textsc{All} (0.81 F). This suggests that lexical information alone can successfully distinguish the CEFR level of course book texts at the document level. Using the complete feature set we obtained 81\% accuracy and 97\% \textit{adjacent accuracy} (when misclassifications to adjacent classes are considered correct). The same scores with lexical features (\textsc{Lex}) only were 80.3\% (accuracy) and 98\% (adjacent accuracy). 

Accuracy scores using other learning algorithms were significantly lower (see Table \ref{alg_res-table}), therefore, we report only the results of the logistic regression classifier in the subsequent sections.

\begin{table}
\caption{\label{alg_res-table} Accuracy scores (in \%) for other learning algorithms.}
\begin{center}
\begin{tabular}{ccccc}
\hline Type & Nr & Perceptron & SMO & J48 \\ 
\hline
\textsc{Lex} & 11 & \bf 77.4 & 42.1 & 55\\
\textsc{All} & 61 & 62.2 & 52.7 & 50.5 \\
\hline
\end{tabular}
\end{center}
\end{table}

Instead of classification, some readability studies (e.g. \cite{huang2011robust,branco2014rolling}) employed linear regression for this task. For a better comparability, we applied also a linear regression model to our data which yielded a correlation of 0.8 and an RMSE of 0.65.

To make sure that our system was not biased towards the majority classes B1 and B2, we inspected the confusion matrix (Table \ref{cm-table}) after classification using \textsc{All}. We can observe from Table \ref{cm-table} that the system performs better at A1 and C1 levels, where confusion occurred only with adjacent classes. Similar to the findings in \cite{franccois2012ai} for French, classes in the middle of the scale were harder to distinguish. Most misclassifications in our material occurred at A2 level (23\%) followed by B1 and B2 level, (20\% and 17\% respectively). 

\begin{table}
\caption{\label{cm-table} Confusion matrix for feature set \textsc{All} at document level.}
\begin{center}
\begin{tabular}{ccccc|c|c}
\hline
\multicolumn{5}{c}{Predictions} & \multicolumn{2}{c}{} \\
\hline
A1 & A2 & B1 & B2 & \multicolumn{3}{l}{C1} \\
\hline
37 & 12 & 0 & 0 & 0 & A1 & L\\
12 & 121 & 18 & 5 & 1 & A2 & a\\
4 & 11 & 206 & 24 & 13 & B1 & b\\
0 & 5 & 21 & 238 & 24 & B2 & e\\
0 & 0 & 0 & 12 & 103 & C1 & l\\
\hline
\end{tabular}
\end{center}
\end{table}

To establish the external validity of our approach, we tested it on a subset of \textsc{LäSBarT} \cite{heimann2013see}, a corpus of Swedish easy-to-read (ETR) texts previously employed for Swedish L1 readability studies \cite{heimann2013see,falkenjack2013features}. We used 18 fiction texts written for children between ages nine to twelve, half of which belonged to the ETR category and the rest were unsimplified. Our model generalized well to unseen data, it classified all ETR texts as B1 and all ordinary texts as C1 level, thus correctly identifying in all cases the relative difference in complexity between the documents of the two categories. 

Although a direct comparison with other studies is difficult because of the target language, the nature of the datasets and the number of classes used, in terms of absolute numbers, our model achieves comparable performance with the state-of-the-art systems for English\cite{heilman2007combining,salesky-shen:2014:W14-18}. Other studies for non-English languages using CEFR levels include: \cite{franccois2012ai}, reporting 49.1\% accuracy for a French system distinguishing six classes; and \cite{branco2014rolling} achieving 29.7\% accuracy on a smaller Portuguese dataset with five levels.

\subsection{Sentence-Level Experiments}
\label{ssec:sent}
After building good classification models at document level, we explored the usability of our approach at the sentence level. Sentences are particularly useful in Computer-Assisted Language Learning (CALL) applications, among others, for generating sentence-based multiple choice exercises, e.g. \cite{VOLODINA14.892}, or vocabulary examples \cite{segler2007investigating}. Furthermore, multi-class readability classification of sentence-level material intended for second language learners has not been previously investigated in the literature.

With the same methodology (section \ref{sec:setup}) and feature set (section \ref{sect:features}) used at the document level, we trained and tested classification models based on the sentence-level data (see section \ref{sect:data}). The results are shown in Table \ref{cl_res_sents-table}.

\begin{table}
\caption{\label{cl_res_sents-table} Sentence-level classification results. }
\begin{center}
\begin{tabular}{ccccc}
\hline Type & Nr & Acc (\%) & F & RMSE \\ 
\hline
\textsc{Majority} & - & 40.2 & 0.23 & 0.49 \\
\textsc{LIX} & 1 &  41.4 & 0.3 & 0.38 \\
\hline
\textsc{Lex} & 11 & 56.8 & 0.53 & 0.33 \\
\textsc{All} & 61 & \bf 63.4 & \bf 0.63 & 0.31 \\
\hline
\end{tabular}
\end{center}
\end{table}

Although the majority baseline in the case of sentences was 7\% higher than the one for texts (Table \ref{cl_res-table}), the classification accuracy for sentences using all features was only 63.4\%. This is a considerable drop (-18\%) in performance compared to the document level (81.3\% accuracy). It is possible that the features did not capture differences between the sentences because the amount of context is more limited on the fine-grained level. It is interesting to note that, although there was no substantial performance difference between \textsc{Lex} and \textsc{All} at a document level, the model with all the features performed 7\% better at sentence level. 

Most misclassifications occurred, however, within a distance of one class only, thus the adjacent accuracy of the sentence-level model was still high, 92\% (see Table \ref{cm_sent-table}). Predictions were noticeably more accurate for classes A1, A2 and B1 which had a larger number of instances.

\begin{table}
\caption{\label{cm_sent-table} Confusion matrix for feature set \textsc{All} at sentence level. }
\begin{center}
\begin{tabular}{ccccc|c|c}
\hline
\multicolumn{5}{c}{Predictions} & \multicolumn{2}{c}{} \\
\hline
A1 & A2 & B1 & B2 & \multicolumn{3}{l}{C1} \\
\hline
371 & 123 & 9 & 2 & 0 & A1 & L\\
120 & 541 & 78 & 11 & 4 & A2 & a\\
27 & 136 & 212 & 23 & 10 & B1 & b\\
8 & 34 & 39 & 30 & 13 & B2 & e\\
0 & 18 & 21 & 9 & 35 & C1 & l\\
\hline
\end{tabular}
\end{center}
\end{table}

\begin{figure}
\centering
\includegraphics[trim = 15mm 5mm 25mm 12cm, clip, width=0.8\textwidth]{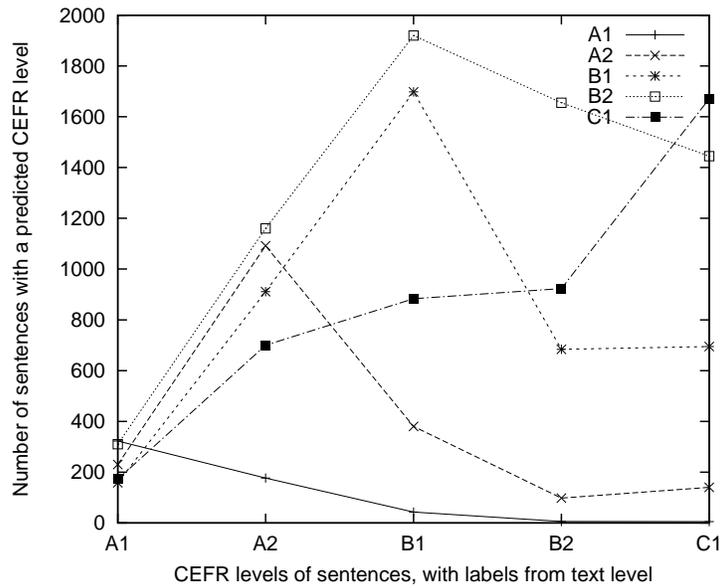}
\caption{Distribution of sentences per CEFR level in the document-level data.}
\label{img:sents_in_texts}
\end{figure}

In the next step, we applied the sentence-level model on the document-level data to explore how homogeneous texts were in terms of the CEFR level of the sentences they contained. Figure \ref{img:sents_in_texts} shows that texts at each CEFR level contain a substantial amount of sentences of the same level of the whole text, but they also include sentences classified as belonging to other CEFR levels.

Finally, as in the case of the document-level analysis, we tested our sentence-level model also on an independent dataset (\textsc{SenRead}), a small corpus of sentences with gold-standard CEFR annotation. This data was created during a user-based evaluation study \cite{Pilan-Ildiko2013-9} and it consists of 196 sentences from generic corpora, i.e. originally not L2 learner-focused corpora, rated as being suitable at B1 or being at a level higher than B1. We used this corpus along with the judgments of the three participating teachers. Since \textsc{SenRead} had only two categories - $<=B1$ and $>B1$, we combined the model's predictions into two classes - A1, A2, B1 were considered as $<=$B1 and B2, C1 were considered as $>$B1. 
The majority baseline for the dataset was 65\%, $<=$B1 being the class with most instances. The model trained on COCTAILL sentences predicted with 73\% accuracy teachers' judgments, an 8\% improvement over the majority baseline. There was a considerable difference between the precision score of the two classes, which was 85.4\% for $<=$B1, and only 48.5\% for $>$B1.

Previously published results on sentence-level data include \cite{Vajjala.Meurers-14-eacl}, who report 66\% accuracy for a binary classification task for English and \cite{dell2014assessing} who obtained an accuracy between 78.9\% and 83.7\% for Italian binary class data using different kinds of datasets. Neither of these studies, however, had a non-native speaker focus. \cite{pilan-volodina-johansson:2014:W14-18} report 71\% accuracy for Swedish binary sentence-level classification from an L2 point of view. Both the adjacent accuracy of our sentence-level model (92\%) and the accuracy score obtained with that model on \textsc{SenRead} (73\%) improve on that score. It is also worth mentioning that the labels in the dataset from \cite{pilan-volodina-johansson:2014:W14-18} were based on the assumption that all sentences in a text belong to the same difficulty level which, being an approximation (as also Figure \ref{img:sents_in_texts} shows), introduced some noise in that data.

Although more analysis would be needed to refine the sentence-level model, our current results indicate that a rich feature set that considers multiple linguistic dimensions may result in an improved performance. In the future, the dataset could be expanded with more gold-standard sentences, which may improve accuracy. Furthermore, an interesting direction to pursue would be to verify whether providing finer-grained readability judgments is a more challenging task also for human raters. 

\section{Conclusion and Future Work}
\label{sect:conclusion}
 
We proposed an approach to assess the proficiency (CEFR) level of Swedish L2 course book texts based on a variety of features. Our document-level model, the first for L2 Swedish, achieved an F-score of 0.8, hence, it can reliably distinguish between proficiency levels. Compared to the wide-spread readability measure for Swedish, LIX, we achieved a substantial gain in terms of both accuracy and F-score (46\% and 0.6 higher respectively). The accuracy of the sentence-level model remained lower than that of the text-level model, nevertheless, using the complete feature set the system performed 23\% and 22\% above the majority baseline and LIX respectively. Misclassifications of more than one level did not occur in more than 8\% of sentences, thus, in terms of adjacent accuracy, our sentence-level model improved on previous results for L2 Swedish readability \cite{pilan-volodina-johansson:2014:W14-18}.

Most notably, we have found that taking into consideration multiple linguistic dimensions when assessing linguistic complexity is especially useful for sentence-level analysis. In our experiments, using only word-frequency features was almost as predictive as a combination of all features for the document level, but the latter made more accurate predictions for sentences, resulting in a 7\% difference in accuracy. Besides L2 course book materials, we tested both our document- and sentence-level models also on unseen data with promising results.

In the future, a more detailed investigation is needed to understand the performance drop between document and sentence level. Acquiring more sentence-level annotated data and exploring new features relying on lexical-semantic resources for Swedish would be interesting directions to pursue. Furthermore, we intend to test the utility of this approach in a real-world web application involving language learners and teachers.

\bibliographystyle{splncs}
\bibliography{cicling_2015}
\end{document}